\documentclass{article}
\usepackage{iclr2027_conference,times}

\usepackage{amsmath,amsfonts,bm}

\def\eqref#1{equation~\ref{#1}}

\def\1{\bm{1}}

\DeclareMathAlphabet{\mathsfit}{\encodingdefault}{\sfdefault}{m}{sl}
\SetMathAlphabet{\mathsfit}{bold}{\encodingdefault}{\sfdefault}{bx}{n}

\usepackage{hyperref}
\usepackage{url}
\usepackage{amsmath,amssymb,bm}
\usepackage{booktabs}
\usepackage{graphicx}
\usepackage{capt-of}
\usepackage{microtype}
\usepackage{xcolor}
\hypersetup{colorlinks=true,linkcolor=black,citecolor=black,urlcolor=blue}

\title{Hierarchical Frequency-Domain Compression of Implicit Geometric Representations for Large-Scale Point Clouds}
\author{Manlin Yao, Jiabin Liu, Guan Wang, Haixu Liu, Hui Li}

\newcommand{\ind}{\mathbf{1}}
\DeclareMathOperator{\round}{round}
\DeclareMathOperator{\clip}{clip}

\iclrfinalcopy
\begin{document}
\maketitle
\fancyhead{}

\begin{abstract}
Large-scale point cloud representations of complex geometries incur prohibitive computational and memory costs, necessitating compressed implicit representations. To address this, we propose a unified framework comprising implicit geometric field representation, hierarchical frequency-domain compression, and conditional high-frequency prediction. Specifically, an unordered point cloud is mapped to an implicit field defined within its physical bounding box. A smooth Fourier pyramid is then constructed, where compact low-frequency components capture the global geometry. Inter-scale high-frequency residuals are encoded to preserve the spatial information required for reconstructing fine geometric details. To restore the high-frequency information lost during compression, we develop a hierarchical 3D neural network. The reconstructed implicit field is converted back into a point cloud through isosurface extraction. Experiments on a complex-boundary point cloud with more than eight million points demonstrate that the proposed method achieves a higher compression ratio than existing point cloud compression methods while maintaining comparable reconstruction quality.
\end{abstract}

\section{Introduction}

Large-scale point clouds provide detailed descriptions of complex objects and three-dimensional scenes, yet their unordered, sparse, and nonuniform spatial distributions make geometry expensive to store and transmit \citep{guo2021survey}. At very low bitrates, a codec must preserve not only the global silhouette but also edges, thin parts, and abrupt curvature changes, which creates a particularly demanding rate--distortion trade-off.

Conventional geometry codecs organize point clouds through octrees, voxel occupancy, or two-dimensional projections \citep{mekuria2017codec,garcia2020geometry,graziosi2020overview}. Learned codecs further reduce redundancy by exploiting sparse convolutions and contextual probability models. Nevertheless, a high-resolution geometry represented voxel by voxel remains costly, and a limited bit budget cannot easily retain global structure and local boundaries at the same time.

Implicit geometric fields map an unordered point set to a regular scalar domain, providing a common representation for continuous surfaces and multiresolution reconstruction \citep{xie2022neuralfields}. Coordinate networks, adaptive spatial decompositions, and multiresolution encodings have demonstrated the expressive power of field-based representations \citep{martel2021acorn,mildenhall2022nerf,muller2022instant}. Their spectral organization also offers a useful separation of roles: low frequencies carry the dominant shape, whereas higher frequencies concentrate boundary transitions and local detail. This observation motivates a compact code built from a stable low-frequency base and a small amount of structured high-frequency evidence.

We introduce a hierarchical frequency-domain compression framework for large-scale point clouds. The encoder first constructs a blockwise implicit geometric field and forms a Fourier analysis pyramid with nested frequency support, retaining only the quantized coarsest field. The residual not represented after upsampling this field is projected onto six fixed low-order 3D DCT modes and quantized with a shared scale, yielding a 29-bit directional descriptor (DCT29). At the decoder, a lightweight Transformer aggregates the compressed descriptions of a target block and its neighbors. The decoded low-frequency base and DCT29 hint are resized to each reconstruction scale, where a conditional decoder progressively restores features and predicts the highest-resolution field.

Our contributions are threefold:
\begin{itemize}
    \item We construct a blockwise implicit geometric field and a Fourier analysis pyramid whose quantized coarsest field provides a deterministic geometric base for reconstruction.
    \item We introduce DCT29, a compact descriptor that stores six fixed directional modes and one shared scale without transmitting frequency indices.
    \item We develop a neighborhood-conditioned multiresolution decoder that converts the shared spectral code into scale-matched bases and directional hints, and uses cross-block context to recover high-resolution geometry.
\end{itemize}

\begin{figure}[t]
    \centering
    \includegraphics[width=\linewidth]{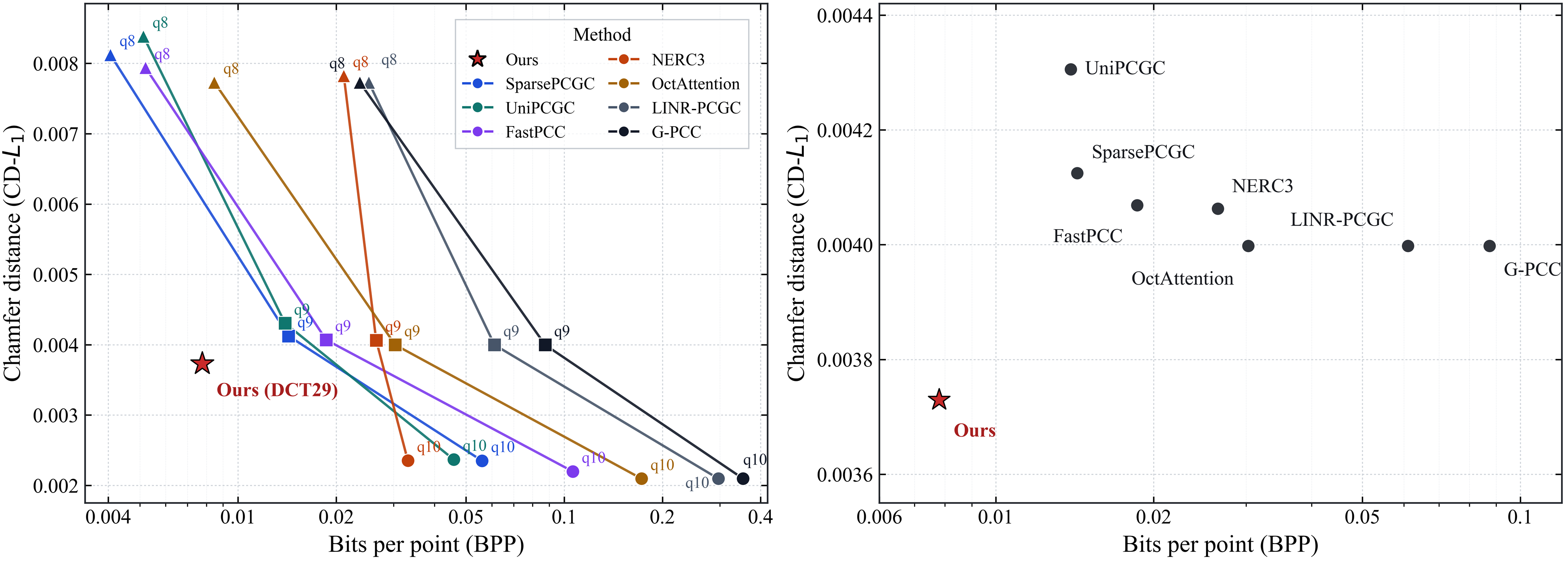}
    \caption{Rate--distortion comparison on the vehicle point cloud. The left panel shows the BPP--CD-$L_1$ operating points of the baselines under q8--q10 quantization; the right panel enlarges the low-rate region. The star denotes the measured configuration of our method.}
    \label{fig:rd_curve}
\end{figure}

\section{Related Work}

\subsection{Conventional and learned point-cloud geometry coding}

Traditional geometry compression relies primarily on spatial subdivision or planar projection. Octree-based codecs describe occupancy through recursive spatial partitioning \citep{mekuria2017codec,garcia2020geometry}. G-PCC combines octree structures with geometry prediction, whereas V-PCC projects the point cloud into two-dimensional atlases and reuses video codecs \citep{schwarz2019mpeg,graziosi2020overview}. Projection padding and surface coding have further improved V-PCC efficiency \citep{li2021padding,xiong2023surface}. Octree methods remain sensitive to the number of occupancy symbols, while projection methods must additionally handle visibility, occlusion, and patch organization.

Learned methods improve the predictability of discrete geometry symbols through probabilistic context models. SparsePCGC models voxel context with sparse convolutions \citep{wang2023sparse}; OctAttention captures long-range dependencies in octree sequences \citep{fu2022octattention}; and UniPCGC unifies multiple multiscale coding modes \citep{wang2025unipcgc}. Recent work has also explored end-to-end lossy geometry coding, hierarchical distortion learning, and inter-frame prediction for dynamic point clouds \citep{zhang2025deeppcc,yu2025fastpcc,akhtar2024interframe}. These approaches operate mainly on occupancy states or learned latent variables. In contrast, we compress a continuous blockwise field and combine an analytic low-frequency base with learned detail recovery.

\subsection{Implicit geometry and neural compression}

Implicit representations describe surfaces with occupancy, distance, or related continuous fields, and can be viewed broadly as mappings from coordinates to field values \citep{xie2022neuralfields}. For geometry compression, NVFPCC \citep{hu2022nvfpcc}, LINR-PCGC \citep{huang2025linr}, NeCGS \citep{ren2025necgs}, and SIR-SNC \citep{wang2026sirsnc} use local latents, sparse features, or scene-specific networks to encode geometry and recover it from a continuous representation.

Such methods typically store most of their information in latent variables, feature grids, or network parameters, so rate depends strongly on model capacity and entropy coding. More generally, nonlinear transform coding shows that the analysis transform, quantization, and probability model jointly determine rate--distortion behavior \citep{balle2021nonlinear}. Our analytic transform instead produces an explicitly quantized low-frequency base and fixed-length directional side information, leaving the neural network responsible only for conditional detail reconstruction.

\subsection{Multiscale spectral representations and conditional reconstruction}

Multiscale transforms organize signals through predefined bases or recursive decompositions. The DCT concentrates energy in an orthogonal cosine basis \citep{ahmed1974dct}, while the Laplacian pyramid separates coarse structure and detail through low-pass filtering, downsampling, and inter-level differences \citep{burt1983laplacian}. For irregular point sets, RAHT constructs a region-adaptive transform from the octree hierarchy \citep{dequeiroz2016raht}; graph transforms define spectral bases from adjacency structure \citep{shuman2013gsp,sandryhaila2013dspg,ortega2018gsp} and have been applied to dynamic point-cloud compression \citep{thanou2016graph}. These methods establish the foundations for quantizing and reconstructing frequency-organized geometry.

Frequency-aware neural representations use encodings, periodic activations, and multiresolution structures to improve high-frequency expressiveness \citep{xie2022neuralfields}. Coarse-to-fine decoders then recover local structure through upsampling and residual prediction. FiLM maps a condition to channel-wise affine parameters \citep{perez2018film}, while self-attention aggregates spatially related information \citep{vaswani2017attention}. Unlike schemes that explicitly transmit multiple transform bands, our codec projects the highest-resolution residual onto six fixed low-order modes known to both encoder and decoder. It transmits only their quantized coefficients and logarithmic scale. During decoding, the same compact code yields scale-matched low-frequency bases and directional hints; neighborhood attention supplies cross-block context, and FiLM modulates reconstruction at each scale.

\section{Method}
\label{sec:method}

\subsection{Problem formulation and framework}

Let $\mathcal{P}=\{\mathbf{p}_i\}_{i=1}^{N}\subset\mathbb{R}^{3}$ be an input point cloud with axis-aligned bounding box $\Omega\subset\mathbb{R}^{3}$. We partition $\Omega$ into regular blocks indexed by $\mathcal{B}$ and place $n_0$ samples along each axis of every block. For block $b\in\mathcal{B}$, the sampling set is
\begin{equation}
\mathcal{X}_b=\{\mathbf{x}_{b,\mathbf{u}}:\mathbf{u}\in\mathcal{I}_{n_0}^{3}\},
\qquad \mathcal{I}_{n}=\{0,\ldots,n-1\}.
\label{eq:samples}
\end{equation}
Let $G:\Omega\rightarrow[0,1]$ denote the implicit field defined below. A block is encoded when its maximum sampled response exceeds an occupancy threshold $\varepsilon_c$:
\begin{equation}
\mathcal{B}_{c}=\left\{b\in\mathcal{B}:\max_{\mathbf{x}\in\mathcal{X}_b}G(\mathbf{x})>\varepsilon_c\right\}.
\label{eq:coded_blocks}
\end{equation}
Each $b\in\mathcal{B}_c$ has quantized symbols $\mathcal{Q}_b$ for its coarse field, DCT coefficients, and shared scale. For neighborhood radius $h$, let $\mathcal{D}_h$ contain the integer offsets whose coordinate magnitudes do not exceed $h$. Neighbor validity is
\begin{equation}
v_{b,\bm{\delta}}=\ind[b+\bm{\delta}\in\mathcal{B}_c],
\qquad \bm{\delta}\in\mathcal{D}_h.
\label{eq:validity}
\end{equation}
Positions outside the bounding box or outside $\mathcal{B}_c$ are zero padded and excluded from attention keys and values. The decoder reconstructs a local field from the target code, valid neighbors, and the normalized global block coordinate $\bm{\pi}_b$:
\begin{equation}
\widehat{G}_b=\mathcal{D}_{\theta}\!\left(
\{\mathcal{Q}_{b+\bm{\delta}},v_{b,\bm{\delta}}\}_{\bm{\delta}\in\mathcal{D}_h},
\bm{\pi}_b\right).
\label{eq:decoder}
\end{equation}

\begin{figure*}[t]
    \centering
    \includegraphics[width=0.70\textwidth]{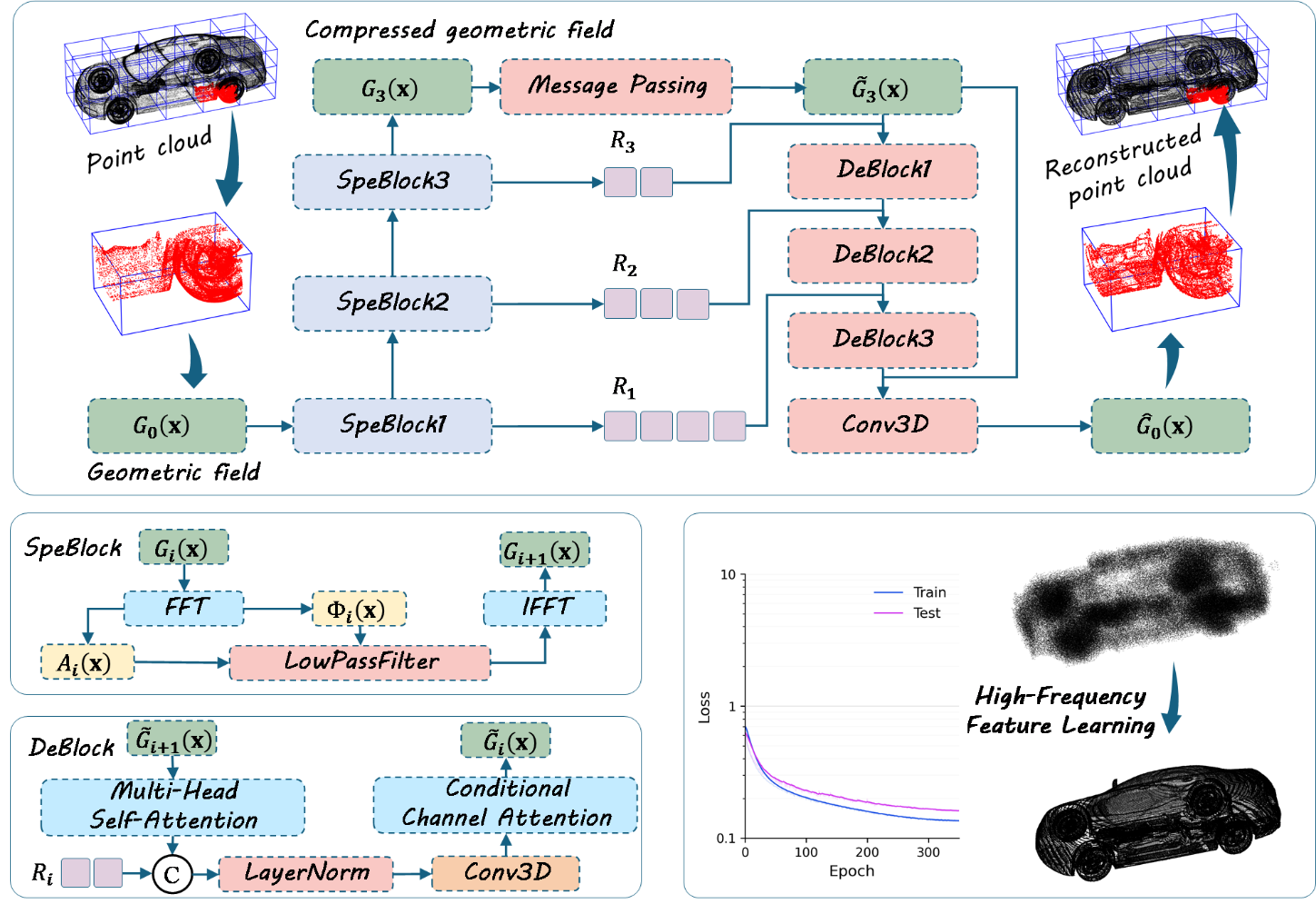}
    \caption{Overview of the proposed codec: blockwise implicit-field construction, hierarchical spectral encoding, neighborhood-conditioned multiresolution decoding, and isosurface recovery.}
    \label{fig:framework}
\end{figure*}

\subsection{Hierarchical frequency-domain block representation}

\paragraph{Sampled implicit geometric field.}
For a grid point $\mathbf{x}\in\Omega$, let $\mathcal{P}(\mathbf{x})\subset\mathcal{P}$ be the input points in its local support. The normalized response of $\mathbf{p}_i\in\mathcal{P}(\mathbf{x})$ is
\begin{equation}
r_i(\mathbf{x})=\exp\!\left[-q\sum_{a\in\{x,y,z\}}
\left(\frac{x_a-p_{i,a}}{L_a}\right)^2\right],
\label{eq:response}
\end{equation}
where $L_a$ is the bounding-box extent along axis $a$ and $q>0$ controls spatial decay. Local responses are aggregated as
\begin{equation}
G(\mathbf{x})=1-\prod_{\mathbf{p}_i\in\mathcal{P}(\mathbf{x})}
\left[1-r_i(\mathbf{x})\right].
\label{eq:field}
\end{equation}
The empty product is defined so that $G(\mathbf{x})=0$ when $\mathcal{P}(\mathbf{x})=\varnothing$; otherwise $0<G(\mathbf{x})\leq1$. Sampling $G$ on $\mathcal{X}_b$ yields the finest block field $G_b^{(0)}$.

\paragraph{Hierarchical Fourier pyramid.}
Let $n_0>n_1>\cdots>n_L=2$ denote the resolutions from fine to coarse. For even $n$, define the centered 3D frequency support
\begin{equation}
\mathcal{K}_n=\left\{-\frac{n}{2},\ldots,\frac{n}{2}-1\right\}^{3}.
\label{eq:freq_support}
\end{equation}
For block $b$ at level $\ell$, the normalized 3D FFT is
\begin{align}
\widehat{G}_b^{(\ell)}(\mathbf{k})
&=\frac{1}{n_\ell^3}\sum_{\mathbf{u}\in\mathcal{I}_{n_\ell}^{3}}
G_b^{(\ell)}(\mathbf{u})
\exp\!\left(-2\pi i\frac{\mathbf{k}^{\mathsf T}\mathbf{u}}{n_\ell}\right) \nonumber\\
&=A_b^{(\ell)}(\mathbf{k})\exp\!\left[i\Phi_b^{(\ell)}(\mathbf{k})\right],
\qquad \mathbf{k}\in\mathcal{K}_{n_\ell}.
\label{eq:dft}
\end{align}
Here $A_b^{(\ell)}$ and $\Phi_b^{(\ell)}$ denote the magnitude and phase, respectively. Let $\mathcal{C}_{\ell}$ be a centered cropping operator that extracts the frequency support associated with resolution $n_{\ell+1}$. Applying the same support to the magnitude and phase gives the cropped complex spectrum
\begin{equation}
\widehat{G}_b^{(\ell+1)}=
\mathcal{C}_{\ell}\!\left(A_b^{(\ell)}\right)
\exp\!\left[i\,\mathcal{C}_{\ell}\!\left(\Phi_b^{(\ell)}\right)\right].
\label{eq:pyramid_spectrum}
\end{equation}
The next-scale field is then obtained by an inverse transform on the target grid:
\begin{equation}
G_b^{(\ell+1)}=
\operatorname{Re}\!\left[
\mathcal{F}_{n_{\ell+1}}^{-1}\!\left(
\mathcal{C}_{\ell}\!\left(A_b^{(\ell)}\right)
\exp\!\left[i\,\mathcal{C}_{\ell}\!\left(\Phi_b^{(\ell)}\right)\right]
\right)\right].
\label{eq:pyramid_inverse}
\end{equation}
The shared support preserves the magnitude--phase correspondence and conjugate symmetry of the retained frequencies, so the inverse transform remains real valued. Repeating this centered low-frequency restriction forms the Fourier pyramid that carries the dominant geometry; compact residual cues and conditional decoding recover the discarded detail.

Only the coarsest field is stored. Its uniform quantization is
\begin{equation}
Q_b^c=\round\!\left[(2^{b_c}-1)G_b^{(L)}\right],
\qquad \overline{G}_b^c=\frac{Q_b^c}{2^{b_c}-1},
\label{eq:coarse_quant}
\end{equation}
where $b_c$ is the number of bits per coarse-field sample. Fourier resampling $\overline{G}_b^c$ to resolution $n_r$ produces the deterministic base $B_b^{(n_r)}$. At the finest resolution,
\begin{equation}
R_b=G_b^{(0)}-B_b^{(n_0)}.
\label{eq:residual}
\end{equation}

\paragraph{Fixed frequency modes and blockwise quantization.}
The residual $R_b$ contains variations absent from the coarse base. Let $\mathcal{J}$ be six fixed low-order 3D DCT-II indices shared by the encoder and decoder: three axial modes and three pairwise coupled modes. With orthonormal separable basis $\Psi_{\mathbf{k}}(\mathbf{u})$, the coefficient of mode $\mathbf{k}$ is
\begin{equation}
c_{b,\mathbf{k}}=\sum_{\mathbf{u}}R_b(\mathbf{u})\Psi_{\mathbf{k}}(\mathbf{u}),
\qquad \mathbf{k}\in\mathcal{J}.
\label{eq:dct_coeff}
\end{equation}
Let $\mathbf{c}_b\in\mathbb{R}^{6}$ collect these coefficients and $q_{\max}$ be the largest magnitude of a signed quantization symbol. The nominal step is set by the maximum coefficient magnitude; the actual symbols use the decoded logarithmically quantized step $\widetilde{\Delta}_b$:
\begin{equation}
\Delta_b=\frac{\lVert\mathbf{c}_b\rVert_\infty}{q_{\max}},
\qquad q_{b,\mathbf{k}}=\clip\!\left(
\round\frac{c_{b,\mathbf{k}}}{\widetilde{\Delta}_b},-q_{\max},q_{\max}\right).
\label{eq:dct_quant}
\end{equation}
The encoder quantizes $\log\Delta_b$ and transmits its scale symbol. The decoded coefficient is $\widetilde{c}_{b,\mathbf{k}}=q_{b,\mathbf{k}}\widetilde{\Delta}_b$; if $\mathbf{c}_b=\mathbf{0}$, all symbols are zero. The finest directional hint is
\begin{equation}
H_b^{(n_0)}=\sum_{\mathbf{k}\in\mathcal{J}}
\widetilde{c}_{b,\mathbf{k}}\Psi_{\mathbf{k}}.
\label{eq:hint}
\end{equation}
It is resampled to every decoder resolution. The occupancy mask, coarse field, and DCT29 descriptors form separate symbol streams and are losslessly entropy coded at the file level.

\subsection{Context-conditioned field reconstruction}

\paragraph{Neighborhood encoding.}
The $3\times3\times3$ neighborhood of a target block is represented by 27 tokens. Each token contains the dequantized coarse field, six DCT coefficients, the shared scale, and a validity flag, together with factorized relative-position embeddings along the three axes. Two multi-head self-attention layers aggregate only valid neighbors to produce a spatial context and a conditioning vector. The normalized global block coordinate supplies additional scene-position information.

\paragraph{Conditional multiresolution decoding.}
The decoder reconstructs features progressively at $4^3$, $8^3$, and $16^3$. Each stage receives its scale-matched low-frequency base, DCT hint, neighborhood context, and the preceding feature map. The frequency descriptor, neighborhood condition, and global coordinate form a block condition $\mathbf{z}_b$, which modulates the feature with FiLM:
\begin{equation}
\widetilde{\mathbf{X}}_b^{(r)}=\mathbf{X}_b^{(r)}\odot
\left(1+\bm{\gamma}_b^{(r)}\right)+\bm{\beta}_b^{(r)},
\quad [\bm{\gamma}_b^{(r)},\bm{\beta}_b^{(r)}]=f_r(\mathbf{z}_b).
\label{eq:film}
\end{equation}
The operations broadcast over spatial locations. The compact DCT hint guides the network toward details not explicitly stored in the bitstream. At the highest resolution, the decoder predicts a bounded correction to the low-frequency base:
\begin{equation}
\widehat{G}_b=B_b^{(n_0)}+\rho\tanh\!\left(
f_{\mathrm{out}}(\mathbf{F}_b^{(n_0)})\right),
\label{eq:output}
\end{equation}
where $\rho>0$ bounds the correction amplitude.

\paragraph{Training objective.}
The objective combines field fidelity, local variation, and block-boundary consistency. For sampled voxel $i$, let $G_i$ and $\widehat{G}_i$ be the target and prediction, $E_i=\widehat{G}_i-G_i$, and $w_i>0$ a regional weight distinguishing the geometric response, its surrounding band, and the background. The field loss is
\begin{equation}
\mathcal{L}_{\mathrm{field}}=
\frac{\sum_i w_i\rho_{\beta}(E_i)}{\sum_i w_i},
\label{eq:field_loss}
\end{equation}
where $\rho_\beta$ is Smooth-$L_1$ with transition parameter $\beta$. Local variation is constrained by matching forward differences along all axes:
\begin{equation}
\mathcal{L}_{\mathrm{grad}}=\frac{1}{3}
\sum_{a\in\{x,y,z\}}\frac{1}{|\Omega_a|}
\sum_{\mathbf{u}\in\Omega_a}
\left|\nabla_a\widehat{G}(\mathbf{u})-\nabla_aG(\mathbf{u})\right|.
\label{eq:grad_loss}
\end{equation}
The boundary term $\mathcal{L}_{\mathrm{boundary}}$ penalizes errors in an intra-block boundary band and discontinuities across adjacent block faces. The complete objective is
\begin{equation}
\mathcal{L}=\mathcal{L}_{\mathrm{field}}+\lambda_g\mathcal{L}_{\mathrm{grad}}+
\lambda_b\mathcal{L}_{\mathrm{boundary}}.
\label{eq:total_loss}
\end{equation}

\subsection{Point-cloud recovery from the implicit field}

The reconstructed blocks are assembled into a global field $\widehat{G}$. Marching Cubes extracts the isosurface at threshold $\tau$ \citep{lorensen1987marching}; a topology-consistent case table can reduce ambiguities \citep{lewiner2003marching}. Vertex $\mathbf{y}_j^{(t)}$ is then refined along the gradient of the trilinearly interpolated field $\widetilde{G}$:
\begin{equation}
\mathbf{y}_j^{(t+1)}=\Pi_{\Omega}\!\left[
\mathbf{y}_j^{(t)}+s_p
\frac{\nabla\widetilde{G}(\mathbf{y}_j^{(t)})}
{\lVert\nabla\widetilde{G}(\mathbf{y}_j^{(t)})\rVert_2+\varepsilon}\right],
\label{eq:vertex_refine}
\end{equation}
where $s_p>0$ is the step size, $\varepsilon>0$ stabilizes normalization, and $\Pi_\Omega$ projects vertices back into the physical bounding box. Nearby duplicates are merged after refinement to obtain the reconstructed point cloud $\widehat{\mathcal{P}}$.

\section{Experiments}
\label{sec:experiments}

\subsection{Experimental setup}

\paragraph{Data and splits.}
The experiments use large-scale point clouds with complex boundaries. The representative vehicle sample contains more than eight million raw points. We construct a global implicit field in each physical bounding box and partition it into regular blocks. Entire HDF5 files are assigned to the training, validation, and test sets with an 80\%/10\%/10\% split, preventing spatial neighborhoods from the same file from crossing subsets. A submarine point cloud provides a supplementary test on a different geometry. Vehicle and submarine error maps use the same one-sided nearest-neighbor definition and a physical-distance color scale.

\paragraph{Baselines.}
We compare with G-PCC \citep{schwarz2019mpeg}, OctAttention \citep{fu2022octattention}, SparsePCGC \citep{wang2023sparse}, UniPCGC \citep{wang2025unipcgc}, FastPCC \citep{yu2025fastpcc}, NeRC$^3$ \citep{ruan2026nerc3}, and LINR-PCGC \citep{huang2025linr}. BPP is computed from the actual compressed file produced by each method, and all reconstructed point clouds are evaluated in a common physical coordinate system.

\paragraph{Metrics.}
For compressed file $F$ with byte length $L_{\mathrm{byte}}(F)$ and input size $N$,
\begin{equation}
\mathrm{BPP}=\frac{8L_{\mathrm{byte}}(F)}{N}.
\label{eq:bpp}
\end{equation}
This rate includes the header, occupancy mask, coarse field, and DCT descriptors. For nonempty reference $\mathcal{P}$ and reconstruction $\widehat{\mathcal{P}}$, geometry error is
\begin{align}
\mathrm{CD}_{L_1}(\mathcal{P},\widehat{\mathcal{P}})=\frac{1}{2}\Bigg[&
\frac{1}{|\mathcal{P}|}\sum_{\mathbf{p}\in\mathcal{P}}
\min_{\widehat{\mathbf{p}}\in\widehat{\mathcal{P}}}
\lVert\mathbf{p}-\widehat{\mathbf{p}}\rVert_2 \nonumber+\frac{1}{|\widehat{\mathcal{P}}|}
\sum_{\widehat{\mathbf{p}}\in\widehat{\mathcal{P}}}
\min_{\mathbf{p}\in\mathcal{P}}
\lVert\widehat{\mathbf{p}}-\mathbf{p}\rVert_2\Bigg].
\label{eq:chamfer}
\end{align}
We also report F-score under a fixed physical-distance threshold and D1-PSNR, both derived from bidirectional nearest-neighbor distances following common point-cloud evaluation practice \citep{tian2017metrics}.

\subsection{Implementation details}

The $2^3$ coarsest field is uniformly quantized to 12 bits. Each of the six DCT coefficients uses a signed 4-bit symbol and shares a 5-bit logarithmic scale, yielding the 29-bit DCT29 descriptor for every coded block. The final BPP is measured from the complete output file, including all symbol streams and required metadata.

The decoder contains approximately 1.79 million parameters. The neighborhood Transformer has two self-attention layers with four heads and 96 channels. The three reconstruction stages use 112, 128, and 96 channels, respectively. We train for 300 epochs with AdamW and cosine learning-rate decay.

\subsection{Rate--distortion performance}

\begin{table}[t]
\caption{Geometry compression results on the large-scale vehicle point cloud. Lower is better for BPP and CD-$L_1$; higher is better for F-score and D1-PSNR.}
\label{tab:main_results}
\centering
\small
\setlength{\tabcolsep}{4.5pt}
\begin{tabular}{lcccc}
\toprule
Method & BPP & CD-$L_1$ ($\times 10^{-3}$)& F-score & D1-PSNR \\
\midrule
OctAttention & 0.0303 & 3.9979& 0.7048 & 61.1256 \\
G-PCC & 0.0874 & 3.9979& 0.7048 & 61.1256 \\
SparsePCGC & 0.0143 & 4.1248& 0.6776 & 60.8562 \\
LINR-PCGC & 0.0611 & 3.9979& 0.7048 & 61.1256 \\
FastPCC & 0.0186 & 4.0686& 0.6895 & 60.9546 \\
NeRC$^3$ & 0.0265 & 4.0627& 0.6934 & 61.0464 \\
UniPCGC & 0.0139 & 4.3054& 0.6513 & 60.5831 \\
\textbf{Ours} & \textbf{0.0078} & \textbf{3.7296}& \textbf{0.7673} & \textbf{60.6035} \\
\bottomrule
\end{tabular}
\end{table}

At 0.0078 BPP, our method achieves a CD-$L_1$ of $3.7296\times10^{-3}$ and an F-score of 0.7673. Relative to UniPCGC, it reduces BPP and CD-$L_1$ by 43.9\% and 13.4\%, respectively, while improving F-score by 17.8\%. Relative to SparsePCGC, it reduces BPP by 45.5\% and CD-$L_1$ by 9.6\%. These results show that hierarchical spectral coding and conditional reconstruction provide a favorable operating point in the low-rate regime. D1-PSNR reaches 60.6035 dB, comparable to the learned baselines. Figure~\ref{fig:rd_curve} shows the measured baseline operating points and highlights the low-rate region containing our configuration.

\subsection{Visual quality and geometric error distribution}

\begin{figure}[t]
    \centering
    \includegraphics[width=\linewidth]{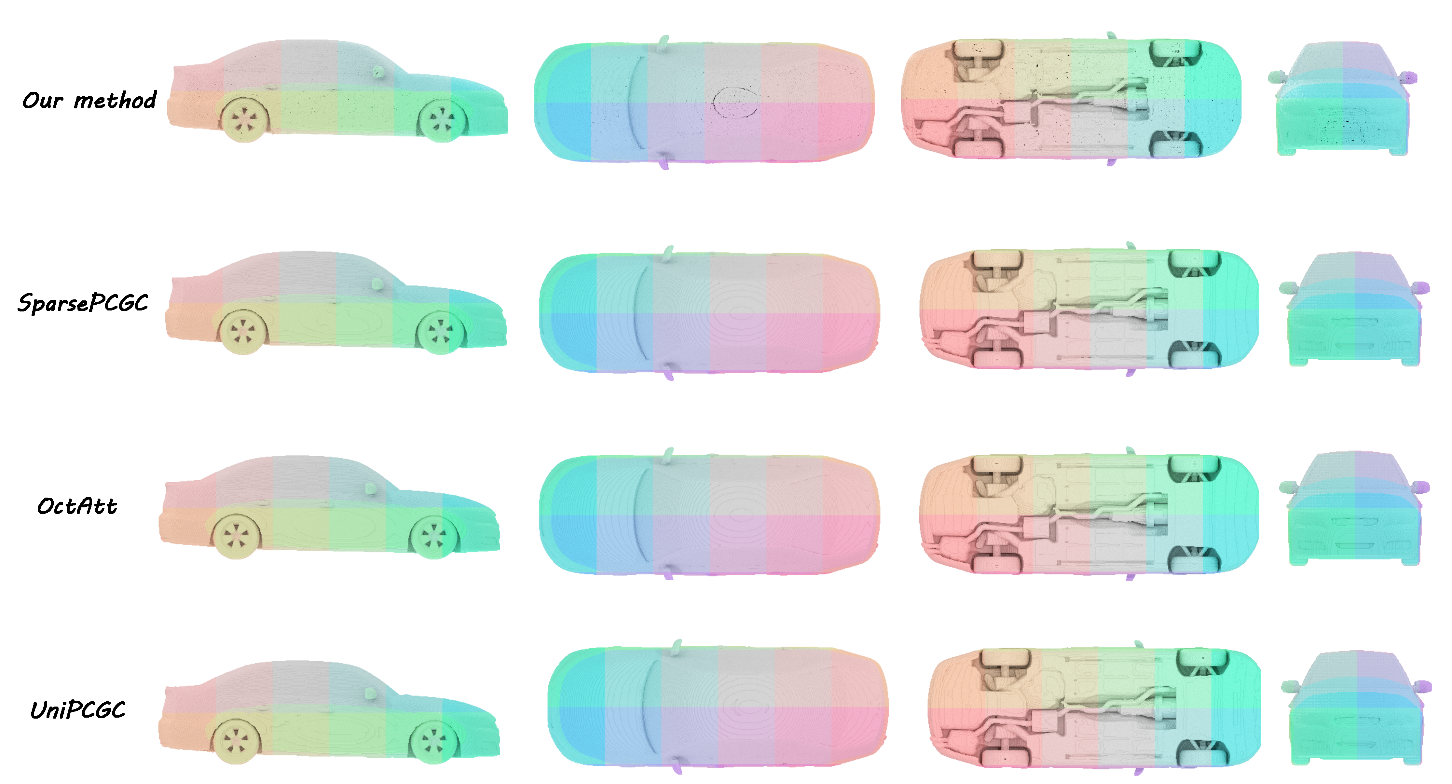}
    \caption{Four-view vehicle reconstructions. Rows show our method, SparsePCGC, OctAttention, and UniPCGC; columns show side, top, bottom, and front views.}
    \label{fig:vehicle_reconstruction}
\end{figure}

Figure~\ref{fig:vehicle_reconstruction} compares the reconstructed vehicle under matched viewpoints. At 0.0078 BPP, our method retains a continuous body outline and recognizable wheels, chassis, and body surfaces, while producing relatively weak banding on smooth regions. Because the methods do not operate at identical rates or reconstruct the same number of points, the figure is used to compare overall shape and local structure; the spatial error maps provide a more direct account of geometric deviation.

For each decoded point $\widehat{\mathbf{p}}\in\widehat{\mathcal{P}}$, its one-sided distance to the reference is
\begin{equation}
d(\widehat{\mathbf{p}},\mathcal{P})=
\min_{\mathbf{p}\in\mathcal{P}}
\lVert\widehat{\mathbf{p}}-\mathbf{p}\rVert_2.
\label{eq:point_error}
\end{equation}

\begin{figure}[t]
    \centering
    \includegraphics[width=0.92\linewidth]{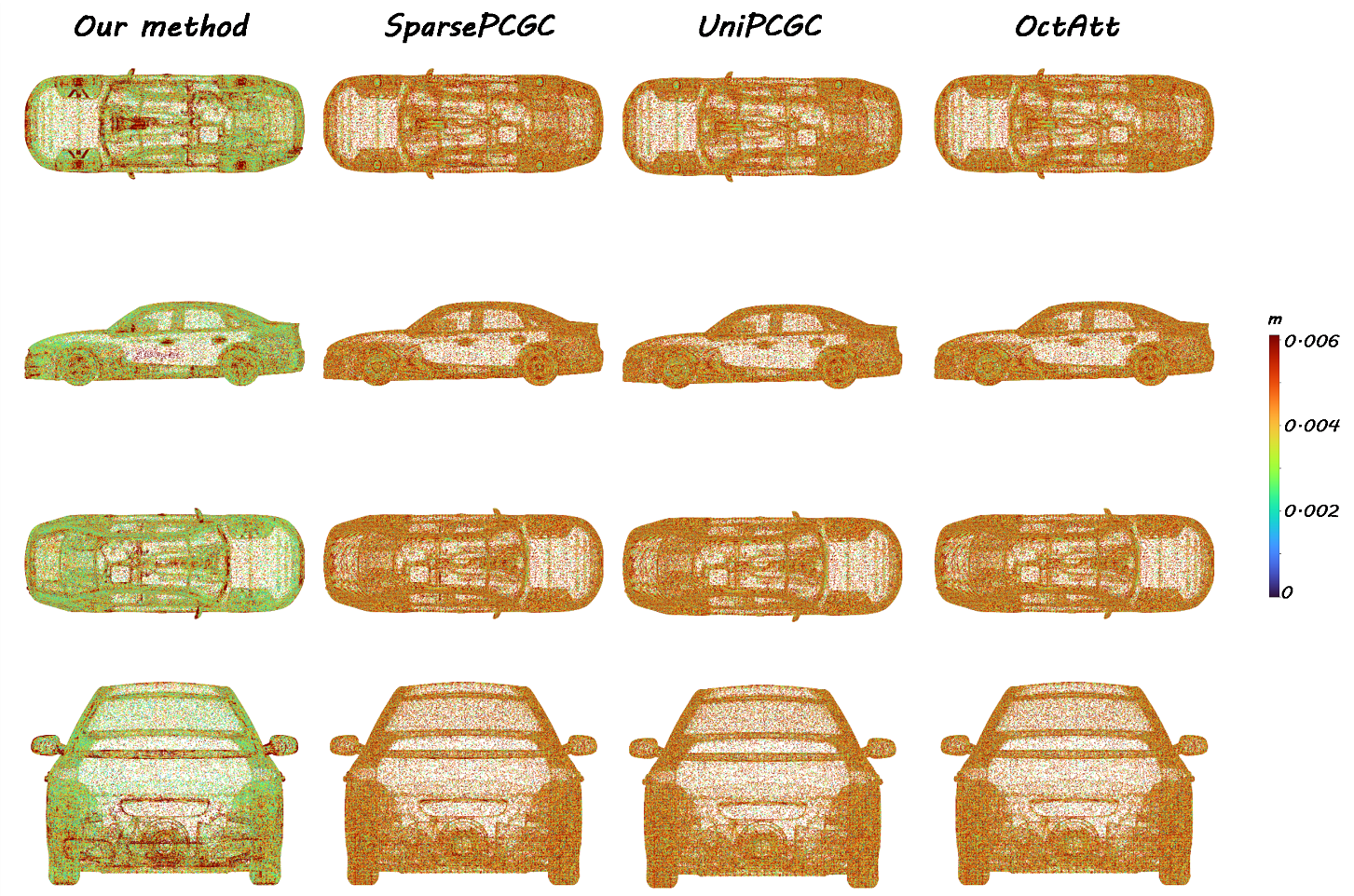}
    \caption{One-sided decoded-to-reference nearest-neighbor error on the vehicle. Columns show our method, SparsePCGC, UniPCGC, and OctAttention; rows show top, side, bottom, and front views. All panels share one physical-distance color scale.}
    \label{fig:vehicle_error}
\end{figure}

Under the shared color scale in Figure~\ref{fig:vehicle_error}, low-error points cover most of the vehicle body reconstructed by our method. Larger deviations are concentrated near the chassis and small local structures. SparsePCGC, UniPCGC, and OctAttention exhibit broader medium- and high-error regions in several views. This spatial pattern is consistent with the CD-$L_1$ values in Table~\ref{tab:main_results}: reconstruction error remains localized rather than producing a global displacement of the main shape.

\begin{figure}[t]
    \centering
    \includegraphics[width=\linewidth]{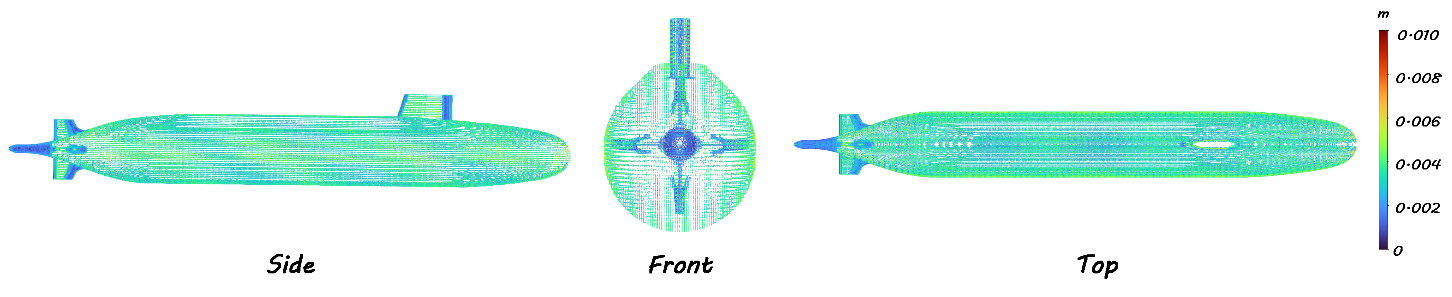}
    \caption{One-sided nearest-neighbor error on the submarine from side, top, and front views. All views use the same physical-distance color scale.}
    \label{fig:submarine_error}
\end{figure}

To examine adaptation to a different geometry, we initialize from the vehicle-trained model and perform a small number of adaptation epochs on the submarine data. Figure~\ref{fig:submarine_error} shows the resulting spatial error distribution. Errors vary smoothly over the hull without large saturated regions and are concentrated near the outer silhouette, conning tower, and tail appendages. The result illustrates reconstruction behavior on an elongated body with local thin structures.

\subsection{Ablation study}

We use the same data split, field construction, isosurface threshold, and evaluation protocol for every ablation. Table~\ref{tab:ablation} compares four settings: Fourier reconstruction from the coarse field only; direct addition of the decoded directional hint to the base; conditional decoding with only the center block; and the complete model with $3\times3\times3$ neighborhood context.

\begin{table}[t]
    \centering
    \caption{Ablation of the spectral representation and conditional reconstruction.}
    \label{tab:ablation}
    \small
    \setlength{\tabcolsep}{2.4pt}
    \begin{tabular}{@{}lccc@{}}
    \toprule
    Setting & BPP & CD-$L_1$ ($\times 10^{-3}$)& D1-PSNR \\
    \midrule
    Low-frequency only& 0.0028 & 38.2999& 38.2361 \\
    LF+HF Encoding& 0.0078 & 41.0933& 38.3369 \\
    Center-block decoder & 0.0078 & 8.5872& 50.5939 \\
    Full model & 0.0078 & 3.7296& 60.6035 \\
    \bottomrule
    \end{tabular}
\end{table}

\begin{center}
    \centering
    \includegraphics[width=0.90\linewidth]{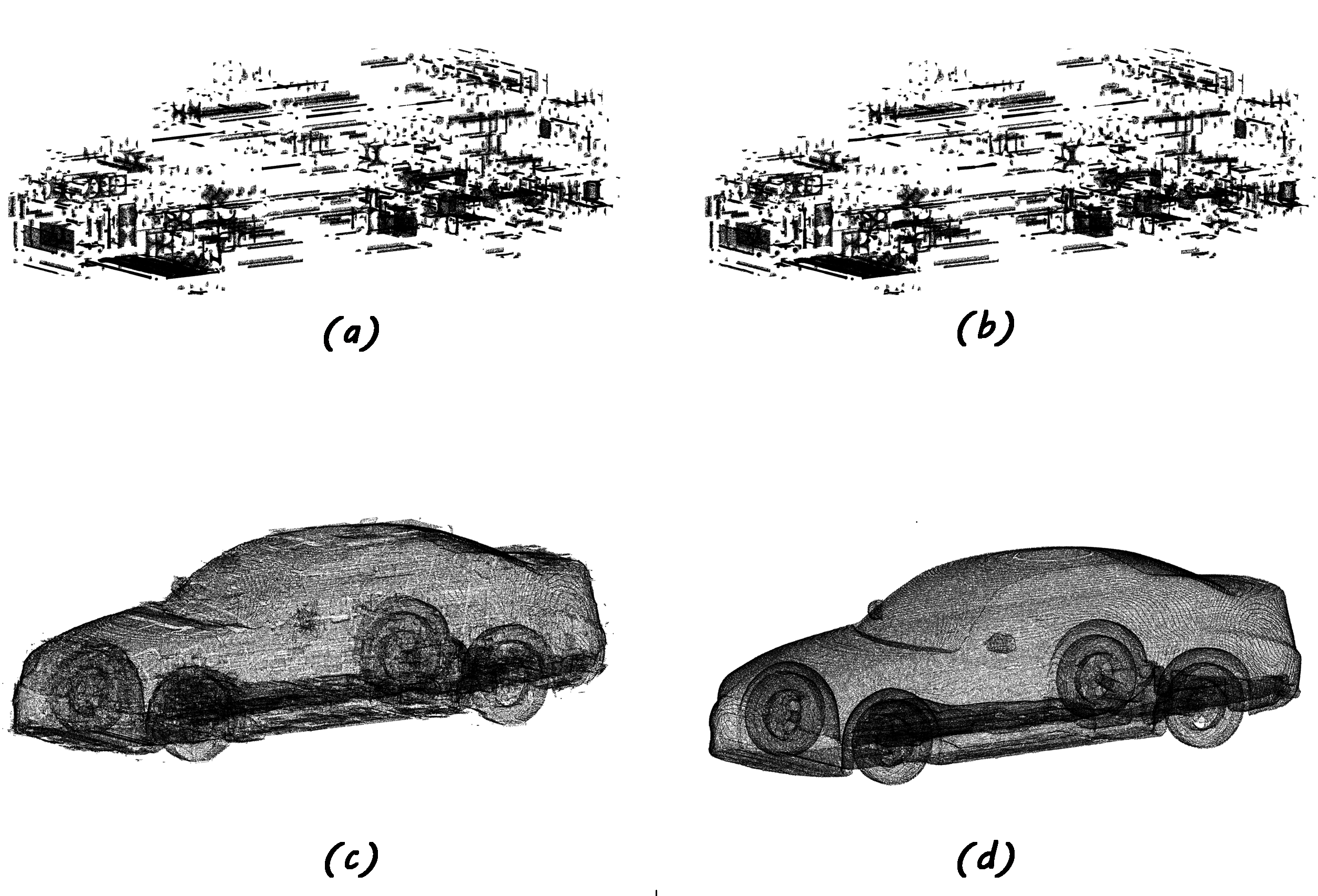}
    \captionof{figure}{Ablation reconstructions: coarse field only, direct DCT compensation, center-block decoder, and full model (left-to-right, top-to-bottom).}
    \label{fig:ablation}
\end{center}

Directly adding the DCT hint increases CD-$L_1$ from $38.2999\times10^{-3}$ to $41.0933\times10^{-3}$ relative to the coarse field, while improving D1-PSNR by only 0.1008 dB. The six fixed modes therefore do not act as a complete residual correction and can introduce local deviations when simply superimposed on the base. Their main value is as conditioning information. At the same 0.0078 BPP, the center-block conditional decoder reduces CD-$L_1$ by 79.1\% and raises D1-PSNR by 12.2570 dB relative to direct compensation, demonstrating that learned reconstruction can make effective use of the compact spectral cue. Adding neighborhood context further lowers CD-$L_1$ from $8.5872\times10^{-3}$ to $3.7296\times10^{-3}$ (56.6\%) and improves D1-PSNR from 50.5939 to 60.6035 dB (10.0096 dB). These gains confirm the importance of adjacent coded blocks for restoring cross-block continuity and refining the highest-resolution field.

\section{Conclusion}

This work presents a hierarchical frequency-domain framework for compressing implicit geometric representations of large-scale point clouds. The proposed codec uses a Fourier analysis pyramid to separate stable coarse geometry from inter-scale detail and projects the residual information onto a compact set of fixed DCT modes shared by the encoder and decoder. Rather than interpreting the resulting coefficients as complete geometric corrections, the decoder uses them as spectral conditions together with the low-frequency base and neighborhood context to reconstruct the field progressively across resolutions. On a large-scale vehicle point-cloud dataset whose representative samples contain more than eight million points, the method achieves a CD-$L_1$ of $3.7296\times10^{-3}$, an F-score of 0.7673, and a D1-PSNR of 60.6035 dB at 0.0078 BPP. Rate--distortion comparisons, spatial error distributions, and controlled ablations consistently demonstrate the complementary roles of the three components: the low-frequency field preserves the dominant geometry, the fixed DCT modes provide compact directional cues, and neighborhood conditioning improves cross-block continuity. These results establish hierarchical spectral conditioning as an effective interface between analytic frequency-domain representation and learned implicit-field reconstruction under a stringent bit budget.

\clearpage
\bibliography{references}
\bibliographystyle{iclr2027_conference}

\end{document}